\documentclass[review]{elsarticle}

\usepackage[english]{algorithm2e}
\usepackage{hyperref}
\usepackage{multirow}
\usepackage{color}
\usepackage{xcolor} 
\usepackage{caption}
\usepackage{subcaption}
\usepackage{balance}
\usepackage{url}
\usepackage{titletoc}
\usepackage{amsmath} 
\usepackage{amssymb}
\usepackage{commath}
\usepackage{lineno,hyperref}
\modulolinenumbers[1]
\usepackage{dirtytalk}
\usepackage{csquotes}
\usepackage{array}
\journal{Journal of Neural Networks}

\biboptions{authoryear}

\begin{document}

\begin{frontmatter}

\title{Adversarial Lagrangian Integrated Contrastive Embedding for Limited Size Datasets}

\author[mymainaddress1]{Amin Jalali}

\author[mymainaddress1,mymainaddress2]{Minho Lee}

\address[mymainaddress1]{KNU-LG Electronics Convergence Research Center, AI Institute of Technology, Kyungpook National University, Daegu, 41566, South Korea}
\address[mymainaddress2]{Graduate School of Artificial Intelligence, Kyungpook National University, Daegu, 41566, South Korea}



\ead{max.jalali@gmail.com, mholee@gmail.com (Corresponding author)}


\begin{abstract}
Certain datasets contain a limited number of samples with highly various styles and complex structures. This study presents a novel adversarial Lagrangian integrated contrastive embedding (ALICE) method for small-sized datasets. First, the accuracy improvement and training convergence of the proposed pre-trained adversarial transfer are shown on various subsets of datasets with few samples. Second, a novel adversarial integrated contrastive model using various augmentation techniques is investigated. The proposed structure considers the input samples with different appearances and generates a superior representation with adversarial transfer contrastive training. Finally, multi-objective augmented Lagrangian multipliers encourage the low-rank and sparsity of the presented adversarial contrastive embedding to adaptively estimate the coefficients of the regularizers automatically to the optimum weights. The sparsity constraint suppresses less representative elements in the feature space. The low-rank constraint eliminates trivial and redundant components and enables superior generalization. The performance of the proposed model is verified by conducting ablation studies by using benchmark datasets for scenarios with small data samples.
\end{abstract}


\end{frontmatter}


\section{Introduction}
\label{sec:intro}
Recent developments in deep learning can be attributed to the increased availability of big data, improvements in hardware and software, and increased speed of the training processes. However, deep neural networks still pose challenges when trained on small datasets. Fine-tuning in small-sample scenarios is prone to overfitting because fitting the model on large-scale parameters with a small number of samples is ill-posed. The limited amount of data samples impedes the use of deep structures in a wide range of applications because achieving good performance requires a large dataset \citep{jiang2022transferability,jalali2020high}. Moreover, the fine-tuning of a model by using a limited datasets can cause overfitting in the downstream target task, particularly when a distributional data gap exists between the pre-trained model and the target task \citep{aghajanyan2020intrinsic,jalali2017sensitive,keisham2022online}. It is also explored that by increasing the model size and training time, the performance of the target tasks gradually saturates to a fixed output \citep{abnar2021exploring}, and in some cases, the performance of the target task is at odds with the pre-trained models. \cite{chen2019catastrophic} noted that when a sufficient number of training samples is available, the spectral components with tiny singular values disappear during the fine-tuning process, implying that small singular values correlate to undesirable pre-trained transfer and may result in a negative transfer. In the fine-tuning process with a supervised objective, self-tuning \citep{wang2021self} can be used to introduce a pseudo-group contrastive approach for evaluating the intrinsic structure of the target domain. Contrastive learning (CL) can be employed to develop generalizable visual features with superior efficiency on target tasks by fine-tuning a classifier on the top of model representations. \cite{fan2021does} analyzed CL in terms of robustness improvement, demonstrating that high-frequency contrastive visual components and the use of feature clustering are advantageous to model robustness without compromising the accuracy. 

In this study, we propose a novel augmented Lagrangian tuning method for an adversarial integrated contrastive embedding model. Adversarially trained models usually have lower accuracy than those trained using the natural training paradigm. However, they perform better when employed for transfer learning to downstream tasks because they generate richer features. We attempt to generate a high feature representation by considering adversarial examples as additional data samples in the training process and do not intend to increase the model robustness against the adversarial examples. The adversarial contrastive mechanism has better transferable knowledge that enables better generalization of small samples. The Lagrangian multiplier intends to adaptively tune the sparsity, low-rank, and accuracy coefficients of the model. 

We summarize our main contributions as follows:
\begin{itemize}
    \item We propose a novel adversarial Lagrangian integrated contrastive embedding called $ALICE$ for small-sized datasets and show the effectiveness of the proposed model on benchmark datasets, namely, CIFAR100, CIFAR10, SVHN, Aircraft, Pets, and Nancho. We introduce the Lagrangian algorithm in the context of adversarial contrastive embedding (AdvCont) to adaptively estimate the coefficients of the constraints.    
    \item We show that the proposed adversarial transfer representation affects downstream target datasets by improving the accuracy and converging faster than the standard training process on different subsets of the datasets with fewer samples.
    \item We present the novel adversarial integrated contrastive model with various augmentation techniques that are fine-tuned with sparsity and low-rank constraint regularizations. The sparsity constraint suppresses less representative elements in the feature space. The low-rank constraint eliminates the trivial and redundant components and facilitates good generalization.  
\end{itemize}

The remainder of this paper is organized as follows. Related works are explained in Section \ref{sec:Related_works}. The proposed ALICE method is described in Section \ref{sec:Lag_adv_con}. The experiments are detailed in Section \ref{sec:experiments}. Finally, the conclusions are presented in Section \ref{sec:conclusion}.

\section{Related Works} \label{sec:related}
\label{sec:Related_works}
In this section, related works regarding transfer learning on small datasets, contrastive learning, and adversarial learning are explained. 

\subsection{Transfer learning on small datasets}
Transfer learning is a technique for achieving high performance in a range of tasks with limited training data. Variant styles, size variability, and shortage of class samples \citep{jalali2019atrial,jalali2021low} make it challenging to achieve good recognition of small-sized datasets. Instead of discovering completely different representations, fine-tuning makes greater use of existing internal representations \citep{li2020rifle}. The model layers have different transfer abilities. The first layers contain general features, middle layers consist of semantic features, and final layers comprise task-specific features. Therefore, different layers should be treated based on the knowledge that is retrained to appropriately fit a target task. Specifically, the first- and mid-layer knowledge should be retained, whereas the last-layer knowledge is adapted to the downstream target tasks. L2 norm with starting point (L2-SP) method utilized $L_2$ constraint regularization to explicitly favor the final target weights to be similar to the pre-trained features \citep{xuhong2018explicit,jalali2015convolutional}. L2-SP demonstrated the effectiveness of establishing an explicit inductive bias towards the original model, and it suggested the $L_2$ regularization penalty term for transfer learning, with the pre-trained model functioning as a baseline model. Mix \& Match \citep{zhan2018mix} employed proxy tasks that were able to generate discriminative representations in the target domain tasks. They developed a mixed process that sparsely selects and combines patches from the target domain to generate diversified features from local patch attributes. Subsequently, a matching process constructed a class-wise linked graph, which helped derive a triplet discriminative objective function to fine-tune the network. To prevent negative transfer, batch spectral shrinkage (BSS) method \citep{chen2019catastrophic} penalized lower singular values to decrease non-transferable spectral components. BSS is a regularization method that suppresses and shrinks non-transferable spectral elements for better fine-tuning. DELTA \citep{li2020delta} presented an attention mechanism to select more discriminative features such that the distance between the features of the pre-trained and target model is regularized. DELTA attempted to preserve higher transferability from a source to a target task. During the fine-tuning process, RIFLE (Re-initializing the fully connected Layer) \citep{li2020rifle} allowed in-depth back-propagation in the transfer learning context by regularly re-initializing the fully connected layers with randomized weight values. RIFLE used an explicit algorithmic regularization strategy to improve low-level feature learning and the accuracy of deep transfer learning. Bi-tuning \citep{zhong2020bituning} is a method for fine-tuning that incorporates two heads into the backbone of pre-trained models. A projector head with a categorical contrastive loss is used to utilize the intrinsic structure of the data samples, and a classifier head with a contrastive loss function is used to consider label information in a contrasting manner. HEAD2TOE \citep{evci2022head2toe} investigated selecting valuable intermediate features from all levels of a pre-trained structure. This approach achieved superior transfer performance when the target affinity differed from the source domain, meaning that the distribution shift was high, and the source-target domain overlap was low. 

\subsection{Contrastive learning}
Compared with the standard training process, contrastive training resulted in better performance on downstream classification tasks \citep{khosla2021supervised}. The standard supervised pre-training transferred high-level feature representations, whereas contrastive counterpart transferred low- and mid-level feature representations. When the downstream task was different from the pre-trained task, the standard pre-training approach had the risk of overfitting high-level features that damaged the transferability \citep{zhao2021makes}. However, contrastive pre-training obtained more generalizable and well-rounded features that obtained better transferability for a variety of downstream tasks. Chen et al. \citep{chen2020simple} introduced CL that maximized the agreement between various augmented views of the same input sample in the latent feature space to generate higher semantic and structural latent representations. \cite{wang2022contrastive} proposed CL with stronger augmentations that used distributional divergence for fine-grained representations from strongly augmented samples. This led to feature representations that were invariant to input disturbances.

\subsection{Adversarial learning}
There was a general perception in adversarial deep learning that robustness and accuracy were mutually conflicting. Recent research has cast doubt on this idea, demonstrating the preservation of robustness while improving the accuracy \citep{zhang2021geometryaware}. They stated that adversarial data samples should be assigned multiple weights. A data point closer to the class border is less robust, and the matching adversarial data point in the feature space should be assigned a larger weight and vice versa. To implement this idea, the weights are determined based on how challenging it is to attack a data point. The experiments showed an increase in the robustness of the standard adversarial training. \cite{utrera2020adversarially} mentioned that adversarial regularization generated representations that retained shapes, lines, and strokes which were more desirable invariable features, while the standard training focused more on obtaining the data textures during the training process. The authors investigated the impact of network architecture and adversarial robustness on different sizes of datasets with adversarial feature fine-tuning on the last convolutional blocks of the model. \citep{salman2020adversarially} also stated that a robust ImageNet classifier resulted in richer feature representations to downstream tasks compared with standard ImageNet counterparts. \cite{wong2020fast} discussed that targeted adversarial attacks such as projected gradient descent (PGD) and Gaussian perturbation obtained significantly better results than a random perturbation. Moreover, a higher number of PGD iterations led to superior transferability. 

In this study, we propose to generate a higher representation by using adversarial examples as additional samples in the training process and employing an adversarially trained model as pre-trained weights for downstream tasks. Moreover, the incorporation of adversarial pre-trained weights in the context of contrastive learning helps to further improve model performance. Finally, bilinear transformation and Lagrangian multipliers are proposed to enforce higher-order representation and classification by considering low-rank and sparsity constraints. This novel approach is presented to achieve improved the accuracy and faster convergence during the training process when using small-sized datasets.

\section{Proposed ALICE}
\label{sec:Lag_adv_con}

\begin{figure}[ht]
\centering\includegraphics[width=0.8\linewidth]{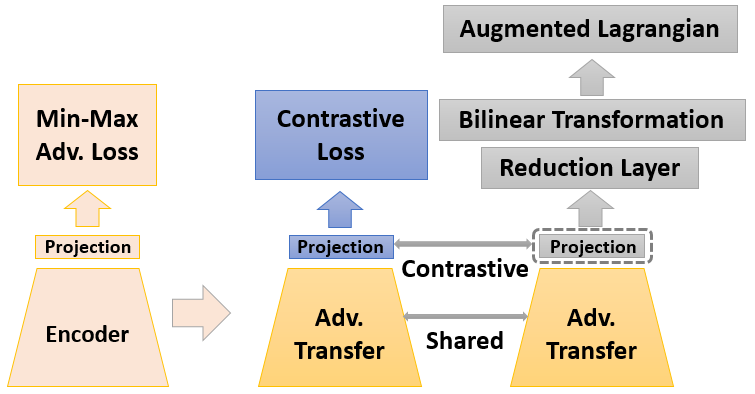}
\caption{Proposed adversarial Lagrangian integrated contrastive embedding (ALICE).}~\label{fig:Lagmodel}
\end{figure}

In this section, we propose a novel adversarial integrated contrastive embedding method to generate model features and fine-tune it with Lagrangian multipliers to encourage low-rank and sparsity improvements. Fig. \ref{fig:Lagmodel} shows the proposed model that fits datasets with limited data samples, which is an error-prone task. First, the images are fed into the left encoder (orange encoder) to train the model using the Min-Max adversarial loss function. Adversarially trained models perform better when employed for transfer learning to downstream tasks because they contain richer features. The adversarial weights obtained from the left encoder are utilized to initialize the middle and right encoders. The images are then fed into contrastive encoders (middle and right yellow encoders) with shared adversarial weights to further train the model with contrastive loss to generate higher representative features on small datasets. Next, we consider only the right encoder as the final model and fine-tune it using the augmented Lagrangian method to adaptively encourage low-rank and sparsity in the model. The gray projection layer is eliminated in the Lagrangian fine-tuning because the reduction layer and bilinear transformation layer are replaced.

\subsection{Preliminaries}
\paragraph{\textbf{Adversarial training}} It produces more diverse samples which lead to better representation. Note that the aim is not to design robust models against adversarial attacks. We intend to generate better feature embedding by considering adversarial examples as additional samples to train the model. The proposed approach generates adversarial instances to maximize loss in a batch. AdvProp \citep{xie2020adversarial} is a method that aligns the statistics of the embedding for both domains of clean and adversarial samples such that they can both contribute to the training of the network. The rationale for generating more diverse instances is to encourage loss to learn more invariant representations.

The adversarial example, $x^{adv}$, of a clean sample, $x$, is defined by Eq. \eqref{eq:adv} for a given learned classifier with a vector of $W=\{w_i\}_{i=1}^C$ parameters. $C$ represents the number of classes in the dataset. $\delta$ denotes adversarial perturbation. $z$ is the classifier feature. Eq. \eqref{eq:adv} shows that the classifier recognizes an image as a different class by adding perturbation to an image and manipulating it. Therefore, in adversarial training, we must ensure that the perturbed image is close to the original image in the feature space and avoid the network being easily fooled by manipulation.
\begin{align}
    \label{eq:adv}
    x^{adv} = x+ \delta \hspace{10pt} \textrm{s.t.}  \hspace{6pt} \operatorname*{argmax}_i w_i^{T}z(x) \neq \operatorname*{argmax}_i w_i^{T}z_(x^{adv})  
\end{align}

Eq. \eqref{eq:minmax} shows the adversarial training objective function in which the maximization loss generates the samples with a perturbation $\delta$ first, and then the minimization loss is performed over the average maximization loss across all $m$ samples. By accommodating the maximization loss of the adversarial regularization in the model minimization loss, we can obtain the Min-Max optimization loss \citep{madry2019deep} as expressed by Eq. \eqref{eq:minmax}. $(x_i, y_i)$ denote the data samples and their corresponding labels. The model is represented by $f_{\theta}(\cdot)$, where $\theta$ is the learnable parameter. The main objective is to minimize the $L(\hat{y}, y)$, where $\hat{y}=f_{\theta}(x)$ is the output prediction score. Adversarial training replaces the minimization loss with Min-Max loss to induce perturbations in the input samples such that the model becomes more robust.  

\begin{equation}
    \label{eq:minmax}
    \operatorname*{min}_{\theta} \frac{1}{m}\sum_{i=1}^{m}  \operatorname*{max}_{{||\delta_i||}_p\leq\epsilon} L(f_{\theta}(x_i,+{\delta}_i), y_i) 
\end{equation}

By solving Eq. \eqref{eq:minmax} over $\theta$, we can obtain the gradient with respect to the parameters of model ${\nabla}_{\theta}$ and update $\theta$ according to Eq. \eqref{eq:theta_update}.  

\begin{equation}
\label{eq:theta_update}
    \begin{split}
    \theta = \theta - \alpha \sum_i \nabla_{\theta} \operatorname*{max}_{{||\delta_i||}_p\leq\epsilon} L(f_{\theta}(x_i,+{\delta}_i), y_i) 
    \end{split}
\end{equation}
$\|\cdot\|_p$ represents the $L_p$ norm, which measures the distance between a generated adversarial example and its clean counterpart. The goal is to find adversarial examples within a distance smaller than the defined threshold $\epsilon$. The optimal amount of perturbation for $x$ is obtained by maximizing the cross-entropy loss in Eq. \eqref{eq:opt_delta} to find the optimal perturbation $\delta^{*}$ that creates the most diverse samples.
\begin{equation}\label{eq:opt_delta}
\delta^{*} = \operatorname*{argmax}_{\delta} L(f_{\theta}(x_i,+{\delta}_i), y_i)  \hspace{10pt} \textrm{s.t.} \hspace{10pt}  \|\delta\|_p <\epsilon
\end{equation}
The training process is performed sequentially, and the worst possible perturbation, ${\delta}_i^{*}$, is calculated for each training sample, $(x_i, y_i)$, based on Eq. \eqref{eq:opt_delta} before updating the model parameters $\theta$. Eq. \eqref{eq:opt_delta} is solved by using projected gradient descent with $k$ update steps called $PGD(k)$ \citep{madry2019deep}. 

We utilize adversarially trained model as transfer learning for the downstream target domain to not only obtain higher accuracy, but also yield faster convergence than standard pre-trained models. The adversarial transfer representation contains rich features as it is trained against adversarial attacks and retains the shapes and more low-level characteristics of the sample images. Next, we explain CL.

\paragraph{\textbf{Contrastive learning (CL)}} It treats each sample as a class and attempts to learn the invariant sample representation. CL generates a pair per sample and aims to pull the samples belonging to the same class together in the embedding space, while simultaneously pushing apart the samples belonging to different classes. CL presents the idea of pulling an anchor sample and a positive sample closer together in the feature embedding space and pushing the anchor apart from many other negative samples \citep{chen2020simple}. As no label information is available in unsupervised learning, a positive pair consists of augmented versions of the sample itself, and the negative pairs are formed by the anchor and randomly chosen samples from the current batch.

We aim to leverage the self-supervised CL approach in a supervised learning setting by utilizing label information. Self-supervised CL contrasts a single positive sample for each anchor against a set of negative samples in the current batch to train the feature embedding space. Supervised CL contrasts the set of all positive samples from a class against all negative samples from other classes in the current batch \citep{khosla2021supervised}. The label information incorporated in the CL generates better representations in the feature embedding space such that samples of the same class are closer to each other. We intend to build {\it many positives} and {\it many negatives} for each anchor using label information, as opposed to self-supervised learning, which employs only a single positive sample. Triplet loss \citep{weinberger2009distance} is closely related to supervised CL, which employs only one positive and one negative per anchor to form pairs for CL, whereas $N$-pair loss \citep{sohn2016improved} employs one positive and many negatives per anchor. In this study, supervised contrastive loss uses many positives and negatives for each anchor, and hard negative mining is not required. 

Given a batch of data samples, $X$, two copies of the batch are generated, $\Tilde{X}$, using augmentation methods $\Tilde{X}=Aug(X)$. Both batches are forwarded through encoder network $Enc(\cdot)$ to extract 2048-D embedding vectors. $Enc(\cdot)$ maps the augmented samples, $\Tilde{X}$, onto the representation vectors $R=Enc(W, \Tilde{X})$. $W$ denotes the weight of the encoder. $R$ is normalized to a hypersphere and then propagated through projection network $z=Proj(R)$ in which the supervised contrastive loss is applied to the output of the projection network. $Proj(\cdot)$ is a multilayer perceptron  with a hidden layer of size 2048 and an output vector of size 128. For a set of $N$ data samples, the number of augmented samples is $2N$. For arbitrary augmented samples $i$ and $j$ belonging to the same class, the self-supervised CL \citep{chen2020simple} method is presented in Eq. \eqref{eq:selfsupervised}. 
\begin{align}
\label{eq:selfsupervised}
    \begin{split}
    &L_{self}(X) =-\sum_{i=1}^{2N} log \frac{exp (z_i \cdot z_j/\tau)}{\sum_{k=1}^{2N} 1_{[k \neq i]} exp (z_i \cdot z_k/\tau) } \\
    &z(X) = Proj(Enc(Aug(X))) \hspace{6pt}\\
    &X=\{x_1,x_2,...,x_i,..., x_{2N}\}
    \end{split}
\end{align}
where \enquote{$\cdot$} calculates the similarity between two vectors. $\tau$ is the temperature parameter and $1_{[k \neq i]} \in \{0,1\}$ returns a value of 1 if $k \neq i$. 
$L_{self}(\cdot)$ calculates the loss across all positive pairs, including both $(i,j)$ and $(j,i)$ in a batch. Given the positive pair $(i,j)$, the other $2(N-1)$ augmented data samples $(i,k)$ are considered to be negatives within the current batch. 

\subsection{Proposed augmented Lagrangian for adversarial contrastive embedding }
\paragraph{\textbf{Proposed AdvCont}} Owing to the presence of label information in CL, Eq. \eqref{eq:selfsupervised} is modified to Eq. \eqref{eq:supervisedcon} to incorporate many positive pairs into the training process. $P(i)$ denotes the set of indices for all positive pairs, and $|P(i)|$ represents the total number of positive pairs. Moreover, the encoders' shared weights, $Enc(W^{Adv}, \Tilde{X})$, are initialized with adversarial pre-trained weights $W^{Adv}$ obtained from the previous stage. $L_{AdvCont}$ calculates the adversarial contrastive learning loss and encourages the encoder to learn the features of samples of the same class. $z^{A}(\cdot)$ denotes the projection network output based on adversarial pre-trained weights. 
\begin{align}
\label{eq:supervisedcon}
    \begin{split}
    &L_{AdvCont}(X) =  \\
    &-\sum_{i=1}^{2N} \frac{1}{|P(i)|} \sum_{p \in P(i)}^{} log \frac{exp (z_i^{A} \cdot z_p^{A}/\tau)}{\sum_{k=1}^{2N} 1_{[k \neq i]} exp (z_i^{A} \cdot z_k^{A}/\tau) }   \\
    &z^{A}(X) = Proj(Enc(W^{Adv}, Aug(X))) 
    \end{split}
\end{align}

In this study, various data augmentation $Aug(\cdot)$ methods are investigated to train the encoder. Mixup \citep{zhang2018mixup}, AutoAug \citep{cubuk2019autoaugment}, and stacked RandAugment \citep{tian2020makes} are selected for our experiments owing to their good performance. The encoder is the ResNet-50 model. Normalized activations of the last pooling layer with 2048-D are employed as the representation vector. The purpose of obtaining better representation is to employ it as a transfer feature for further fine-tuning. 

Temperature parameter $\tau$ plays a crucial role in improving the performance of the model. Given the positive $(p)$, negative $(n)$, and anchor $(a)$ samples in Eq. \eqref{eq:triplet}, it is shown that $L_{Cont}$ is a simple case of triplet loss with only one positive and one negative sample that creates the margin of $\alpha=2\tau$ between classes \citep{weinberger2009distance}. We empirically determined that the $\tau$ value to be 0.1 for optimal accuracy.  
\begin{align}
\label{eq:triplet}
    \begin{split}
        L_{Cont} &=-log {\frac{exp (z_a^{A} \cdot z_p^{A}/\tau)}{exp (z_a^{A} \cdot z_p^{A}/\tau)+exp (z_a^{A} \cdot z_n^{A}/\tau)}}\\
        &=log(1+exp((z_a^{A} \cdot z_n^{A} - z_a^{A} \cdot z_p^{A})/ \tau))\\
        &\textrm{By using Taylor expansion of $log$ function:} \\
        &\approx exp((z_a^{A} \cdot z_n^{A} - z_a^{A} \cdot z_p^{A})/ \tau)  \hspace{10pt} 
        \approx 1+ \frac{1}{\tau}(z_a^{A} \cdot z_n^{A} - z_a^{A} \cdot z_p^{A})\\
        &= 1- \frac{1}{2\tau} (\|z_a^{A} -z_n^{A}\|^2 - \|z_a^{A}-z_p^{A}\|^2)
        \propto \|z_a^{A}-z_p^{A}\|^2-\|z_a^{A} -z_n^{A}\|^2 +2\tau
    \end{split}
\end{align} 

An adversarial representation is incorporated into CL to present the proposed $AdvCont$ for the problem of data scarcity. In Section~\ref{sec:experiments}, the effectiveness of the proposed $AdvCont$ with various augmentation strategies is illustrated for encoders on different benchmark datasets. A good representation has a strong inductive bias, which makes $AdvCont$ more data-efficient and leads to better generalization when there is data scarcity. Next, we present bilinear transformation and an augmented Lagrangian algorithm to accommodate sparsity and low-rank regularizations in the structure of $AdvCont$. We fine-tune the $AdvCont$ model to enforce sparsity and low-rank enhancements using multi-objective multipliers.

\paragraph{\textbf{Bilinear transformation for AdvCont}} It is a technique used in image classification to obtain superior feature representation based on higher-order feature extraction between channels \citep{li2018towards}. Compact bilinear pooling (CBP) \citep{gao2016compact} was proposed to approximate bilinear features by reducing the high-dimensional issue. Bilinear features model the second-order information (i.e., covariance matrices) of the input samples that have superior representation and discrimination; however, they present certain challenges such as redundant features, huge computational burden, and overfitting.
Recently, iSQRT \citep{li2018towards} employed the Newton-Schulz iteration (NSI) \citep{higham1997stable}, which is GPU-friendly because it only uses matrix multiplication to approximate matrix square-root normalization. iSQRT cannot accommodate low-rank and sparsity constraints to promote generalization, compactness, or stronger representation \citep{min2020multi}. All the above methods utilized NSI to approximate the square-root ($L_2$ norm) normalization, which was not sufficient to stabilize the high-order statistics of the features well. NSI was strictly based on the calculation of matrix derivatives, which required the function to be continuously differentiable at the neighborhood of the root to perform the matrix inverse operation. However, sparsity and low-rank constraints are not differentiable and have different convex properties to be optimized jointly for $AdvCont$ model.

We intend to normalize the matrix of bilinear transformation for $AdvCont$ in terms of the accuracy, sparsity, and low-rank, as shown in Fig. \ref{fig:Lagmodel}. $L_2$ norm attempts to stabilize second-order bilinear representation. Sparsity suppresses less distinctive elements in high-dimensional feature representations, and low-rank eliminates trivial model structure components. These regularizing terms promote second-order information stabilization, compactness, and generalization. However, these regularizers exhibit different convex properties. Therefore, the augmented Lagrangian method formulates regularizer constraints with auxiliary variables to loosen the correlations between the different regularizing terms. A closed-form solution to solve each constraint alternately is obtained based on the alternating direction method of multipliers (ADMM) \citep{boyd2011distributed}.

\begin{figure}[ht]
\centering\includegraphics[width=1\linewidth]{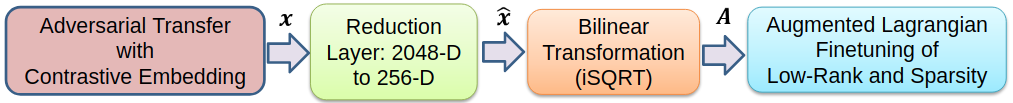}
\caption{Pre-trained adversarial transfer with contrastive embedding, the reduction layer, the bilinear transformation, and the augmented Lagrangian objective function.}~\label{fig:reduc_bilinear}
\end{figure}

Gradient-based multi-objective optimization is based on Karush-Kuhn-Tucker (KKT) conditions \citep{peitz2018gradient} and employed to find a descending direction for all defined objectives. The multi-objective augmented Lagrangian algorithm is used to fine-tune the adversarial transfer with contrastive embeddings (AdvCont) as shown in Fig. \ref{fig:reduc_bilinear}. We remove the projection layer from $AdvCont$ model, as shown in Fig. \ref{fig:Lagmodel} and add a reduction layer containing $(Conv+BN+ReLU)$ to reduce the dimension of second-order statistics bilinear features. The last convolutional layer feature map is defined as $X \in R^{b \times h \times w \times c}$, where $b, h, w$, and $c$ denote the batch size, height, width, and channels of the feature map, respectively. The reduction layer decreases the number of channels $c$ from $X \in R^{b \times h \times w \times 2048}$ to $ \Bar{X} \in R^{b \times h \times w \times 256}$. For simplicity, we reshape $m=h \times w$, where $\widehat X \in R^{b \times m \times 256}$. $\widehat X$ is a matrix containing feature maps of a batch of data. Next, the bilinear features are obtained by calculating the covariance matrix, which is a symmetric positive definite (SPD) matrix as shown in Eq. \eqref{eq:covmat}. Bilinear features from iSQRT contain stronger representation and discrimination characteristics because the bilinear operator models the second-order information of the input samples.
\begin{equation}
\label{eq:covmat}
    A= \widehat X^{T} \big( \frac{1}{n}(I_{n \times n}- \frac{1}{n} 1_{n \times n}                           \big) \widehat X    
\end{equation}
where  $I_{n \times n}$ denotes the identity matrix and $1_{n \times n}$ represents the all-ones matrix. The determinant of the SPD matrix is always greater and equal to zero; therefore, its direction and singular values will not be changed. The bilinear features $(A)$ are fed into the multi-objective augmented Lagrangian algorithm to be fine-tuned based on the square-root, sparsity, and low-rank regularizers.

\begin{figure}[ht]
\centering\includegraphics[width=0.7\linewidth]{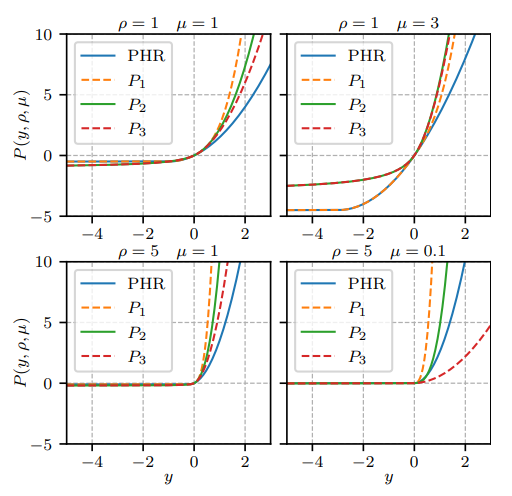}
\caption{Plotted Lagrangian penalty functions \cite{birgin2005numerical} corresponding to Eq. \eqref{eq:penaltyfunc}.}~\label{fig:Lagrangfunc}
\end{figure}

\paragraph{\textbf{Proposed AdvCont with augmented Lagrangian tuning}}
Given a minimization objective function, $g(x)$, with a constraint function, $h(x)$, penalty methods convert the constraint problems in \eqref{eq:objcons} to an unconstrained problem by adding a penalty term. The value of the penalty term increases when the constraint is violated.  
\begin{align} 
\label{eq:objcons}
    \begin{split}
    &min(g(x)) \hspace{10pt} \textrm{s.t.} \hspace{10pt} h(x) < 0  \hspace{50pt} \textrm{Constrained}\\
    &min(g(x)+\alpha h(x))  \hspace{10pt}  g,h : R^d \rightarrow R   \hspace{10pt} \textrm{Unconstrained}
    \end{split}
\end{align}
where $\alpha$ is heuristically determined. The augmented Lagrangian approach (ALA) has the advantage of estimating the $\alpha$ value automatically compared with the penalty method \citep{bertsekas2014constrained}. ALA adaptively estimates the multipliers, which are the optimal weights for the constraints that avoid the ill-conditioning issue in penalty methods. It uses a succession of iterations to solve the optimization problem. The outer iteration is indexed by $i$ and the inner iteration is indexed by the input $x$. During the inner iteration, the augmented Lagrangian function, $G(x)$, in Eq. \eqref{eq:lageq} is minimized w.r.t. $x$. Once the inner convergence criterion is satisfied, the outer iteration, which consists of {\it penalty multiplier} $\mu$ and {\it penalty parameter} $\rho$, is updated. In the outer iteration, $\mu$ is updated to the derivative of $P$ with respect to $h(x)$ in Eq. \eqref{eq:muupdate}.

\begin{equation}
    \label{eq:lageq}
    G(x)=g(x)+P(h(x), \rho^{(i)}, \mu^{(i)})
\end{equation}
\begin{equation}
    \label{eq:muupdate}
    \mu^{(i+1)}= P^{'}(h(x),\rho^{(i)},\mu^{(i)})
\end{equation}
$P(\cdot)$ is a Lagrangian penalty function, such that  $P^{'}(y,\rho,\mu)=\frac{\partial}{\partial{y}}P(y,\rho,\mu)$ exists. Any candidate function, $P(\cdot)$, should satisfy these criteria \citep{birgin2005numerical}: i) the derivative of $P(\cdot)$ should be positive and equal to $\mu$ when $y=0$, and ii) the derivative of $P(\cdot)$ with respect to $y$ tends to zero if the constraint is satisfied $(h(x)<0)$ and tends to infinity if $(h(x)\geq 0)$. This is an adaptive update of the penalty weight, where $\mu$ increases when the constraint is not satisfied; otherwise, $\mu$ decreases. In the outer iteration, the value of $\rho$ gradually increases if the constraint function has not reduced significantly. $\rho$ increases to a higher value to reach the Lagrangian penalty function to its ideal penalty. Eq. \eqref{eq:penaltyfunc} presents the examples of the most common Lagrangian functions \citep{birgin2005numerical, rony2021augmented}, and the corresponding plotted functions with various values of $\mu$ and $\rho$ is depicted in Fig. \ref{fig:Lagrangfunc}. The augmented Lagrangian algorithm is presented in Algorithm \ref{alg:algorithm}.

\begin{align}
    \begin{split}
    \label{eq:penaltyfunc}
    &PHR(y,\rho,\mu)=\frac{1}{2\rho}(max\{0,\mu+\rho y\}^{2}-\mu^{2})\\
    &P_1(y,\rho,\mu)= \begin{cases}
    \mu y + \frac{1}{2}\rho y^{2}+ \rho^{2}y^{3},   & \text{if} \hspace{20pt}  y\geq 0\\
    \mu y+\frac{1}{2}\rho y^{2},                    & \text{if} \hspace{20pt} -\frac{\mu}{\rho} \leq y \leq 0\\
    -\frac{1}{2\rho} \mu^{2},                        & \text{if} \hspace{20pt} y \leq -\frac{\mu}{\rho}
    \end{cases}\\
    &P_2(y,\rho,\mu)= \begin{cases}
    \mu y + \mu\rho y^{2}+ \frac{1}{6}\rho^{2}y^{3},   & \text{if} \hspace{20pt}  y\geq 0\\
    \frac{\mu y}{1-\rho y},                            & \text{if} \hspace{20pt}  y\leq 0 
    \end{cases}\\
    &P_3(y,\rho,\mu)= \begin{cases}
    \mu y + \mu\rho y^{2},                            & \text{if} \hspace{20pt}  y\geq 0\\
    \frac{\mu y}{1-\rho y},                            & \text{if} \hspace{20pt}  y\leq 0 
    \end{cases}
    \end{split}
\end{align}

\begin{algorithm}
\scriptsize \small
\SetAlgoLined
\SetKwInOut{Input}{Input}
\SetKwInOut{Output}{Output}
\SetKwInOut{Init}{Initialization}

\Input{ Loss function $g(.)$, constraint function $h(.)$, penalty function $P(.)$, iteration $i$, batch size $b$,  }
\Output{Minimize the loss with unconstrained function $G(.)$ w.r.t $(x)$}
\Init{Initialize input $x^{(1)}$, penalty multiplier $\mu^{(1)}$, and penalty parameter $\rho^{(1)}$}
\For{i=1 to N}{
	\For{b=1 to B}{
	    Use $x^{(1)}$ as the initialization and minimize $G(x)$
	    $G(x)=g(x)+P(h(x), \rho^{(i)}, \mu^{(i)})$\;
    	$x^{(i+1)} \leftarrow$ approximate minimizer of $G(x)$\;
        $\mu^{(i+1)} \leftarrow$ update $P^{'}(h(x),\rho^{(i)},\mu^{(i)})$\;
        if the constraint does not improve, then
            set $\rho^{(i+1)}>\rho^{(i)}$\;
	}
}
 \caption{Generic augmented Lagrangian method}
 \label{alg:algorithm}
\end{algorithm}

Based on the Lagrangian penalty function $P_{1}(h(x), \rho, \mu)$ defined in Eq. \eqref{eq:penaltyfunc}, and the simpler version \citep{bertsekas2014constrained} of the objective function, $G(x)$, in Eq. \eqref{eq:lageq} is formed as Eq. \eqref{eq:extendedlageq}. The reason for the presence of the $augmented$ term in the augmented Lagrangian algorithm is the presence of an additional quadratic penalty term in Eq. \eqref{eq:extendedlageq} that ensures better convergence speed and superior stabilization. 
\begin{equation}
    \label{eq:extendedlageq}
    G(x)=g(x)+ trace(\mu^{(T)}h(x))+\frac{\rho}{2} \|h(x)\|^{2}
\end{equation}
Based on Eq. \eqref{eq:objcons}, we write the entire objective function including square-root, low-rank, and sparsity as defined in Eq. \eqref{eq:multiobjectives}.

\begin{equation}
\label{eq:multiobjectives}
    \begin{split}
    &min \hspace{4pt} g(x)+\alpha_1 h_1(x)+\alpha_2 h_2(x) \\
    &=\operatorname*{min}_{Y} \|Y^{2}-A\|_F^{2}+\alpha_1\|Y\|_* + \alpha_2 \|Y\|_1
    \end{split}
\end{equation}
where $\|\cdot\|_F$ represents the Frobenius norm that constrains $Y^{2}$ to be similar to the covariance matrix $A$. $\|\cdot\|_{*}$ is the nuclear norm used to approximate the rank function by calculating the sum of singular values of the input matrix. $\|\cdot\|_1$ denotes the $L_1$ norm to encourage sparsity by approximating the $L_0$ norm of the matrix. $Y$ represents the target-regularized features. $\alpha_1$ and $\alpha_2$ are constraint coefficients. This objective function is non-smooth and has various convex properties; therefore, the augmented Lagrangian method is used to loosen the correlations between these three constraints.  Two auxiliary variables, $J_1$ and $J_2$, are defined to reformulate the constraints independently as Eq. \eqref{eq:J_variable}, which is a constrained optimization problem with three loosen penalty terms.
\begin{align} 
\label{eq:J_variable}
    \begin{split}
    &\operatorname*{min}_{Y,J_1,J_2} \|Y^{2}-A\|_F^{2}+\alpha_1\|J_1\|_* + \alpha_2 \|J_2\|_1 \\ &\textrm{s.t.}  \hspace{7pt}  J_1=Y, J_2=Y 
    \end{split}
\end{align}
By using the augmented Lagrangian multiplier introduced in Eq. \eqref{eq:extendedlageq}, we can convert Eq. \eqref{eq:J_variable} and obtain the unconstrained form of the objective function as Eq. \eqref{eq:Lan_objfunc}.
\begin{align} 
\label{eq:Lan_objfunc}
    \begin{split}
    &\operatorname*{min}_{Y,J_1,J_2} \|Y^{2}-A\|_F^{2}  \\
    &+ \alpha_1 \|J_1\|_* + trace(\mu_1^{(T)} (J_1 - Y))+\frac{\rho_1}{2} \|(J_1 - Y)\|^{2} \\
    &+ \alpha_2 \|J_2\|_1 + trace(\mu_2^{(T)} (J_2 - Y))+\frac{\rho_2}{2} \|(J_2 - Y)\|^{2}
    \end{split}
\end{align}
where $trace(\cdot)$ calculates the sum of diagonal elements of the matrix, $\mu_1$ and $\mu_2$ are Lagrange multipliers, and $\rho_1$ and $\rho_2$ are the penalty parameters. The closed-form solution of \eqref{eq:Lan_objfunc} can be solved by using the ADMM \citep{boyd2011distributed} to update each variable alternately. The ADMM first updates $J_1$ while keeping $J_2$ and $Y$ variables fixed. Therefore, irrelevant terms in Eq. \eqref{eq:Lan_objfunc} are eliminated. This process is repeated to update $J_2$ and then $Y$. After the three variables ($J_1, J_2, Y$) are updated, Lagrangian multipliers $\mu_1$ and $\mu_2$ and penalty parameters $\rho_1$ and $\rho_2$ are updated by applying Eq. \eqref{eq:multiplierupdate} for the $i^{th}$ iteration where hyperparameter $m=1.1$ and $\mu_1=\mu_2=10$.
\begin{align} 
\label{eq:multiplierupdate}
    \begin{split}
    & \mu_1^{i+1}= \mu_1^{i}+\rho_1(J_1^{i} - Y^{i})\\
    & \mu_2^{i+1}= \mu_2^{i}+\rho_2(J_2^{i} - Y^{i})\\
    &\rho_1 \leftarrow m \rho_1; \rho_2 \leftarrow m \rho_2
    \end{split}
\end{align}

Therefore, the proposed ALICE model attempts to improve the training convergence and the accuracy performance with adversarial transfer integrated with CL for small datasets. Then, the augmented Lagrangian multipliers adaptively encourage optimum low-rank and sparsity values on the covariance matrix of bilinear features. The second-order covariance matrix of the input samples obtained from the bilinear mechanism provides a better representation \citep{li2018towards} and discrimination. Moreover, the Lagrangian method is proposed because the GPU-friendly iSQRT bilinear transformation cannot accommodate low-rank and sparsity owing to differentiability issues. The calculation of matrix derivatives requires the function to be continuously differentiable in the neighborhood of the root to conduct the matrix inverse operation. Thus, the Lagrangian algorithm formulates all the constraints in a closed-form solution to solve each constraint alternately using the ADMM.

\section{Experiments}
\label{sec:experiments}

We perform the experiments for 200 epochs using adversarial transfer to adapt the last three convolutional blocks of standard ResNet-50 to the small target datasets. We then train the contrastive embedding for 150 epochs. The learning rate throughout these 350 epochs is $\eta_1= 0.01$. In the second stage of training, we fine-tune $AdvCont$ model for 30 more epochs to encourage sparsity, low-rank, and square-root enhancements with a learning rate of $\eta_2= 0.001$. The learning rate in the second stage is lower than that in the first stage ($\eta_2=0.001< \eta_1= 0.01$) for smooth training. The batch size is 128, momentum is equal to 0.9, and weight decay is set to $5 \times 10e-4$. The temperature value is heuristically obtained as $\tau=0.1$. 

$\boldsymbol{Datasets:}$ CIFAR100 and CIFAR10 datasets contain 50,000 samples for the training stage and 10,000 samples for the test stage. CIFAR100 contains 100 classes, whereas CIFAR10 contains 10 classes. SVHN dataset has 73,257 samples for training and 26,032 samples for testing with 10 classes. The Aircraft dataset comprises 6,667 samples for the training process and 3,333 samples for the test process with 100 classes. The Pets dataset comprises 3,680 samples for training and 3,369 for testing with 37 classes. The Nancho dataset contains 6,350 samples for the training stage and 2,822 samples for the test stage with 280 classes.

The Academy of Korean Studies provides a small-sized Nancho dataset \footnote{Dataset download link: https://github.com/AI-repo/Datasets} for research on the translation of ancient cursive Korean archives into modern Korean. The visually similar features make it difficult for the model to distinguish samples with high commonalities \citep{jalali2019atrial}. Documents have image degradation, including document aging and issues with the quality of the ink, such as ink dispersion due to the passage of time. Moreover, in some cases, the documents are of low quality because they date back several hundred years. Because of aging, the ink is dispersed over the edges, making them difficult to read. These highly degraded samples exhibit a lower recognition performance. The samples include all types of disturbances, including ink dispersion, extensive cursive styles, low-quality resolution, and complex backgrounds. The Nancho dataset is a small dataset, and the samples are segmented directly from documents obtained from various source scripts. We perform various experiments on different smaller subsets of the mentioned public datasets to verify the novelty and impact of the proposed ALICE model on small datasets. 

\begin{table}[ht!]
\caption{Performance of standard transfer versus adversarial transfer for 10\%, 30\%, and 100\% of the target dataset to check the performance of the transfer with few samples in the dataset. The accuracy values are mentioned in terms of mean($\pm$Std).}
\label{tbl:fewer_instances}
\setlength{\tabcolsep}{3.5pt}
\begin{center}
\begin{tabular}{|l|c|ccc|}
\hline
    
\textbf{Datasets} &\textbf{Transfer Method}  &     &\textbf{Data Percentage}    &     \\
\textbf{}         &\textbf{}  &\textbf{10\%} &\textbf{30\%}  &\textbf{100\%}  \\ \hline   \hline
                  
CIFAR100      &Standard           &55.8(0.7) &75.0(0.6)&81.5(0.4)   \\ 
              &Adversarial        &60.9(0.5) &76.5(0.4)&81.9(0.2)   \\ \hline
CIFAR10       &Standard           &83.4(0.6) &91.8(0.4)&95.1(0.3)   \\ 
              &Adversarial        &87.5(0.4) &93.0(0.3)&95.7(0.2)   \\ \hline
SVHN          &Standard           &74.0(0.9) &88.6(0.6)&95.3(0.4)   \\ 
              &Adversarial        &84.0(0.6) &92.4(0.4)&96.0(0.3)   \\ \hline

\end{tabular}
\end{center}
\end{table}

When the model is adversarially trained on a large feature-rich dataset such as ImageNet, it captures more information from the shapes, strokes, lines, and patterns of the samples. Owing to the presence of perturbations in adversarial regularization, the model learns robust representations that convey better features. Table \ref{tbl:fewer_instances} shows that adversarial regularization transfers superior representation and semantic information, particularly when the dataset contains fewer training instances. This experiment is conducted by considering a subset of 10\% and 30\% of the target datasets to check the effect of adversarial transfer compared to standard transfer. In this experiment, the pre-trained ImageNet transfer is used for standard transfer learning, whereas the pre-trained adversarial ImageNet transfer with constraint $||\delta||_2 \leq 3$ is used for adversarial transfer. In the adversarial training process, PGD(20) is used to take 20 update steps to obtain the worst perturbation, and the model parameters $\theta$ are then updated by loss minimization. Adversarial transfer generalizes better to target datasets, especially when there are fewer samples. For example, as indicated in Table \ref{tbl:fewer_instances}, for the CIFAR10 dataset when 10\% of the dataset is employed for training, the performance difference between the standard and adversarial transfer is $87.5\%-83.4\%=4.1\%$. However, the difference when the entire dataset (100\%) is employed is $95.7\%-95.1\%=0.6\%$. The adversarially transferred method exhibits better gradients. It improves performance by having lower standard deviation (Std) values. For example, as presented in Table \ref{tbl:fewer_instances}, for the CIFAR100 dataset, the accuracy (mean $\pm$ Std) of the model using 100\% of the dataset with adversarial transfer method is 81.9(0.2), whereas that for the standard transfer method is 81.5(0.4). This shows that the adversarial transfer method has a higher accuracy $(81.9>81.5)$ and lower standard deviation $(0.2<0.4)$.
\begin{table}[ht!]
\caption{Performance of standard transfer versus adversarial transfer at different epochs (40, 60, 100, 150, and 200) for the perturbation ratio of $||\delta||_2 \leq 3$ on the target dataset with a subset having 10\% of the data size. The mean accuracy values are mentioned in the table and the standard deviation values are denoted as Std.}
\label{tbl:faster_epoch}
\setlength{\tabcolsep}{3.5pt}
\begin{center}
\begin{tabular}{|l|c|ccccc|}
\hline
                  
\textbf{Datasets} &\textbf{Transfer }  && &\textbf{Epochs} &&   \\
\textbf{    (10\%)}         &\textbf{Method}  &\textbf{40}   &\textbf{60}     &\textbf{100}   &\textbf{150}  &\textbf{200}     \\ \hline  \hline
                  
CIFAR100 &Standard &47.6(0.8) &53.2(0.8) &55.3(0.7) &55.5(0.7) &55.8(0.7) \\
    &Adversarial   &56.8(0.6) &58.7(0.6) &60.2(0.5) &60.5(0.5) &60.9(0.5) \\ \hline
CIFAR10 &Standard &80.0(0.7) &81.1(0.6) &83.2(0.6) &83.3(0.6) &83.4(0.6)   \\ 
    &Adversarial   &85.4(0.5) &86.2(0.5) &87.1(0.4) &87.2(0.4) &87.5(0.4) \\ \hline
SVHN &Standard  &67.8(0.9) &70.5(0.9) &73.9(0.9) &73.9(0.9) &74.0(0.9)    \\
    &Adversarial &81.9(0.7)  &82.2(0.7) &83.9(0.7) &84.0(0.6) &84.0(0.6)   \\ \hline

\end{tabular}
\end{center}
\end{table}

\begin{figure}[ht!]
\centering\includegraphics[width=0.8\linewidth]{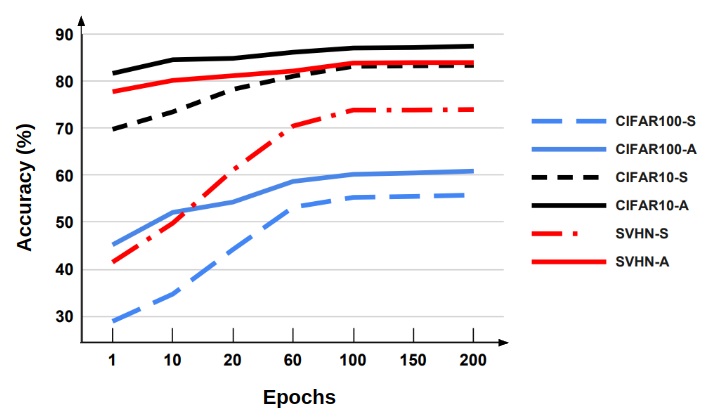}
\caption{The convergence accuracies for both the standard and adversarial methods for various datasets.}~\label{fig:acc-chart}
\end{figure}
The encoder trained with adversarially robust regularization term transfers faster to the target domain than standard training.  Table \ref{tbl:faster_epoch} presents the performance of the standard transfer compared with adversarial transfer using the perturbation value of $||\delta||_2 \leq 3$ on target datasets with 10\% of the dataset. The accuracy is shown at different epochs of 40, 60, 100, 150, and 200 to illustrate that the adversarial transfer is faster than the standard transfer. For example, for the SVHN dataset at $40^{th}$ epoch, the standard method results in $67.8\%$ accuracy, whereas the adversarial fine-tuning approach results in $81.9\%$ accuracy, indicating that adversarial transfer converges faster at fewer epochs. The standard deviation (Std) values of the adversarial transfer method are also lower, indicating better stabilization and less variation.  

Fig. \ref{fig:acc-chart} demonstrates the convergence accuracies for both the standard and adversarial methods after more epochs (i.e., 150 and 200). The standard and adversarial training methods are represented by “S” and “A” in Fig. \ref{fig:acc-chart}. “CIFAR100-S” denotes the training with standard method for “CIFAR100” dataset. The accuracy values of the model for different datasets converge to certain numbers. For example, the convergence value for CIFAR10-S and CIFAR10-A are $83.4\%$ and $87.5\%$.This experiment is the illustration of the convergence mentioned in Table \ref{tbl:faster_epoch}. The experiment considers only 10\% of the data size to show better convergence of adversarial pre-trained initialization in the low-regime data size. For example, for the SVHN dataset at the 60th epoch, the standard method (dashed red, SVHN-S) resulted in 70.5\% accuracy, whereas the adversarial fine-tuning approach (solid red, SVHN-A) resulted in 82.2\% accuracy, indicating that adversarial transfer converges faster with fewer epochs.

\begin{table}[ht!]
\caption{Performance of standard transfer versus adversarial transfer on small datasets for the perturbation ratio of $||\delta||_2 \leq 3$ on the target dataset.}
\label{tbl:small-data-size}
\setlength{\tabcolsep}{3pt}
\begin{center}
\setlength{\tabcolsep}{2pt}
\begin{tabular}{|l|c|c|c|cc|}
\hline

\textbf{Datasets} &\textbf{Size} &\textbf{Classes}   &\textbf{Transfer}            &\textbf{Accuracy}   &\textbf{Accuracy}    \\
                  &\textbf{(Train/Test)}    &    &\textbf{Method}
 &\textbf{ResNet-50}   &\textbf{WRNet-50-2}   \\\hline  \hline
                  
Aircraft    &6,667/3,333    &100    &Standard    &86.1$\pm$0.4 &86.7$\pm$0.4    \\ 
            &               &       &Adversarial &86.3$\pm$0.3 &86.9$\pm$0.3    \\ \hline
Pets     &3,680/3,369    &37     &Standard  &90.8$\pm$0.2       &91.3$\pm$0.2 \\ 
         &               &       &Adversarial  &91.1$\pm$0.1    &91.5$\pm$0.1 \\ \hline
Nancho  &6,350/2,822     &280    &Standard  &76.5$\pm$0.4  &77.1$\pm$0.3   \\ 
        &               &       &Adversarial &78.4$\pm$0.2      &78.9$\pm$0.2                \\ \hline
\end{tabular}
\end{center}
\end{table}

Table \ref{tbl:small-data-size} also presents the performance of adversarial transfer compared with standard transfer on three small datasets, namely, the Aircraft, Pets, and Nancho datasets, with a perturbation ratio of $||\delta||_2 \leq 3$ on the target datasets. The experiments are conducted on two ResNet-50 and WRNet-50-2 backbones. The results indicate that adversarial transfer retains more low- and mid-level patterns that convey more features to the target downstream datasets. WRNet-50-2 backbone also outperforms the ResNet-50 model. 

\begin{table*}[ht!]
\caption{Performance of the adversarial cross-entropy (AdvCE), adversarial contrastive embedding (AdvCont), and added augmentation methods such as Mixup \citep{zhang2018mixup}, Stacked RandAug \citep{tian2020makes}, and AutoAug \citep{cubuk2019autoaugment} on CIFAR100, CIFAR10, SVHN, Aircraft, Pets, and Nancho datasets. The accuracy is written in terms of mean($\pm$ standard deviation).}
\label{tbl:advcont_aug}
\begin{center}
\setlength{\tabcolsep}{5pt}
\begin{tabular}{|l|ccccc|}
\hline
                  
\textbf{Datasets}   &\textbf{AdvCE}  &\textbf{AdvCont} &\textbf{AdvCont}  &\textbf{AdvCont}   &\textbf{AdvCont}    \\                                                   &                &                 &\textbf{Mixup}    &\textbf{Rand-Aug}   &\textbf{AutoAug}    \\ \hline \hline
                  
CIFAR100    &81.9(0.2)&82.6(0.2)&82.9(0.3)&83.2(0.3)&83.6(0.2)   \\   \hline
CIFAR10     &95.7(0.2)&96.3(0.2)&96.6(0.2)&96.8(0.3)&97.1(0.2)   \\   \hline
SVHN        &96.0(0.3)&96.5(0.2)&96.7(0.2)&96.9(0.2)&97.2(0.2)   \\   \hline
Aircraft    &86.3(0.3)&87.2(0.2)&87.6(0.3)&87.9(0.3)&88.1(0.2)   \\   \hline
Pets        &91.1(0.1)&91.6(0.1)&92.0(0.2)&92.2(0.2)&92.3(0.1)   \\  \hline
Nancho      &78.4(0.2)&79.3(0.2)&80.1(0.3)&80.7(0.2)&81.0(0.2)   \\   \hline
\end{tabular}
\end{center}
\end{table*}

Table \ref{tbl:advcont_aug} presents an evaluations of the performance of adversarial cross-entropy (AdvCE), AdvCont, and the integration of AdvCont with different augmenters for various datasets. AdvCE is an adversarially trained model. We used its pre-trained weights for transfer learning to downstream tasks and fine-tuned it using the cross-entropy (CE) loss function. Experiments show that leveraging adversarial transfer with contrastive learning (AdvCont) generalizes better and outperforms the AdvCE approach. AdvCont pulls together samples of the same class, while pushing away samples from different classes, in the embedding space transferred by adversarial training to generate better representation. AdvCont outperforms the AdvCE method by contrasting the samples from the same class as the positive category in the embedding space against the rest of the samples in the batch as the negative category. For example, for the CIFAR100 dataset, the increased performance is $82.6\%-81.9\%=0.7\%$ and for the Nancho dataset, the enhanced accuracy is $79.3\%-78.4\%=0.9\%$. We also analyze the effects of different augmenters, such as Mixup \citep{zhang2018mixup}, Rand-Aug \citep{tian2020makes}, and AutoAug \citep{cubuk2019autoaugment} in the AdvCont model. Mixup is an augmentation method that integrates images with specific probabilities. AutoAug operates with an augmentation policy approach trained by using reinforcement learning. Rand-Aug employs a random parameter tuned by AutoAug to reduce the search space. Table \ref{tbl:advcont_aug} displays the effectiveness of the AutoAug method compared with other augmentation methods. For instance, the results for the SVHN dataset display that $AdvCont+AutoAug$ outperforms the baseline $AdvCont$ by margin of $97.2\%-96.5\%=0.7\%$. In the following experiments, $AdvCont+AutoAug$ is considered as the main backbone for further Lagrangian fine-tuning of sparsity and low-rank combinations because it outperforms other structures.

\begin{table*}[ht!]
\caption{Ablation study of the proposed ALICE method compared with various methods in the literature on benchmark datasets.}
\label{tbl:spar_lowrank_effect}
\begin{center}
\setlength{\tabcolsep}{4pt}

\begin{tabular}{|l|cccccc|}
\hline

\textbf{Models}   &\textbf{C100} &\textbf{C10}  &\textbf{SVHN}  &\textbf{Aircraft}  &\textbf{Pets}  &\textbf{Nancho}\\ \hline  \hline  
 
HEAD2TOE                        &54.6     &NA     &86.3   &NA   &89.1   &NA\\ 
\citep{evci2022head2toe}         &   &   &   &   &   &  \\ \hline
L2SP          &81.4     &95.1     &NA   &86.5   &89.4 &NA\\
\citep{xuhong2018explicit}           &   &   &   &   &   &  \\ \hline
Mix \& Match              &80.6     &95.0     &NA   &87.4   &89.6 &NA\\ 
\citep{zhan2018mix}           &   &   &   &   &   &  \\ \hline
DELTA                 &80.4     &94.7     &NA   &87.0   &89.5 &NA\\ 
\citep{li2020delta}         &   &   &   &   &   &  \\ \hline
BSS          &80.4     &94.8     &NA   &87.2   &89.5 &NA\\ 
\citep{chen2019catastrophic}          &   &   &   &   &   &  \\ \hline
RIFLE                &80.3     &94.7     &NA   &87.6   &90.0 &NA\\ 
\citep{li2020rifle}       &   &   &   &   &   & \\  \hline
SupCon    &81.5     &95.3     &NA   &87.4  &89.7  &NA\\
\citep{gunel2021supervised}           &   &   &   &   &   & \\ \hline
Bi-Tuning           &81.4     &95.1     &NA   &87.4   &89.9 &NA\\ 
\citep{zhong2020bituning}           &   &   &   &   &   &  \\ \hline       \hline 
Proposed AdvCont    &83.6     &97.1     &97.2   &88.1   &92.3   &81.0\\
+AutoAug          &   &   &   &   &   &  \\ \hline
Proposed AdvCont    &84.0     &97.4     &97.6   &88.5   &92.5   &81.5\\ 
+AutoAug+Sparsity             &   &   &   &   &   & \\  \hline
Proposed AdvCont   &84.2     &97.7     &97.8   &89.0   &92.7   &81.9\\
+AutoAug+Low-Rank             &   &   &   &   &   &  \\ \hline
\textbf{Proposed ALICE}     &\textbf{84.9}  &\textbf{98.0}  &\textbf{98.3} &\textbf{89.5}   &\textbf{92.9}   &\textbf{82.4}\\ \hline
  
\end{tabular}
\end{center}
\end{table*}

Table \ref{tbl:spar_lowrank_effect} shows the ablation study of the proposed ALICE compared with various methods in the literature on benchmark datasets, namely, HEAD2TOE \citep{evci2022head2toe}, L2SP \citep{xuhong2018explicit}, Mix \& Match \citep{zhan2018mix}, DELTA \citep{li2020delta}, BSS \citep{chen2019catastrophic}, RIFLE \citep{li2020rifle}, SupCon \citep{gunel2021supervised}, and Bi-tuning \citep{zhong2020bituning}. We conduct the ablation studies, as shown in Table \ref{tbl:spar_lowrank_effect}, to evaluate the components of the Lagrangian regularizers and compare with other methods in the literature. Table \ref{tbl:spar_lowrank_effect} indicates that sparsity incorporation into the $AdvCont+AutoAug$ model results in better performance as it suppresses less effective elements in the feature space. For example, $AdvCont+AutoAug+Sparsity$ outperforms the baseline $AdvCont+AutoAug$ for SVHN dataset by a margin of $97.6\%-97.2\%=0.4\%$. The sparse constraint also enhances the performance of the model for other datasets. The integration of low-rank in $AdvCont+AutoAug$ also eliminates trivial and less representative components, which reduces redundancies and improves the performance. As indicated in Table \ref{tbl:spar_lowrank_effect}, the model $AdvCont+AutoAug+Low-Rank$ achieves better generalization than $AdvCont+AutoAug$ for the Aircraft dataset with a margin of $89.0\%-88.1\%=0.9\%$, which shows the constraint effectiveness. The proposed $ALICE$, which accommodates Lagrangian low-rank and sparsity regularizations in the adversarial integrated contrastive embedding model, exhibits the highest performance on six benchmark datasets compared with other methods. It generalizes better because it jointly optimizes all non-smooth regularizations with various convex characteristics. The margin improvement of the accuracy for $ALICE$ compared with $AdvCont+AutoAug$ for the Nancho dataset is $82.4\%-81.0\%=1.4\%$ and for CIFAR10 is $98.0\%-97.1\%=0.9\%$. 

\begin{table*}[ht!]
\caption{Ablation study of the proposed ALICE method compared with various methods in the literature on benchmark datasets.}
\label{tbl:proposed-wrn}
\begin{center}
\setlength{\tabcolsep}{6pt}

\begin{tabular}{|l|cc|cc|}
\hline

\textbf{Datasets}  &\multicolumn{2}{c|}{\textbf{AdvCont+AutoAug}} & \multicolumn{2}{c|}{\textbf{ALICE}}  \\

&\textbf{ResNet-50}  &\textbf{WRNet-50-2} &\textbf{ResNet-50}  &\textbf{WRNet-50-2}  \\  \hline  \hline  

CIFAR100 &83.6$\pm$0.2   &84.0$\pm$0.3   &84.9$\pm$0.2   &85.2$\pm$0.2   \\ \hline
CAFAR10  &97.1$\pm$0.2   &97.4$\pm$0.2   &98.0$\pm$0.2   &98.2$\pm$0.3   \\ \hline
SVHN     &97.2$\pm$0.2   &97.4$\pm$0.3   &98.3$\pm$0.3   &98.4$\pm$0.2   \\ \hline
Aircraft &88.1$\pm$0.2   &88.2$\pm$0.2   &89.5$\pm$0.2   &89.7$\pm$0.3   \\ \hline
Pets     &92.3$\pm$0.1   &92.7$\pm$0.2   &92.9$\pm$0.2   &93.2$\pm$0.2   \\ \hline
Nancho   &81.0$\pm$0.2   &81.7$\pm$0.3   &82.4$\pm$0.3   &83.0$\pm$0.2   \\ \hline

\end{tabular}
\end{center}
\end{table*}

\begin{figure}[ht!]
\centering\includegraphics[width=0.8\linewidth]{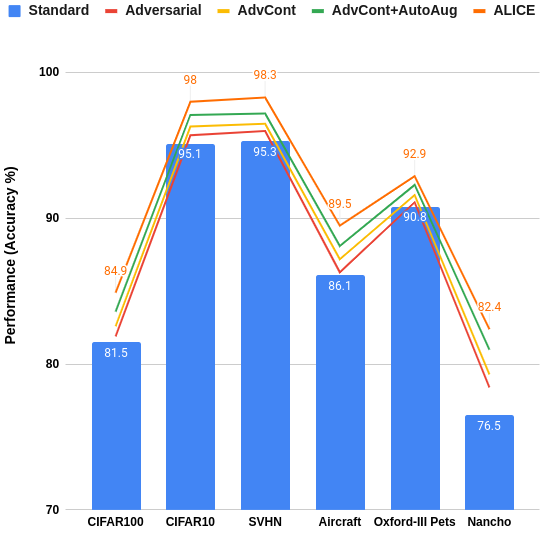}
\caption{Improvement effect of adversarial transfer, AdvCont, AdvCont with AutoAug, and ALICE model on baseline model for various datasets.}~\label{fig:chart}
\end{figure}

Table \ref{tbl:proposed-wrn} shows the performance of the proposed $AdvCont+AutoAug$ and $ALICE$ using the ResNet-50 and WRNet-50-2 backbones. The WRNet-50-2 backbone outperforms the ResNet-50 backbone on the benchmark datasets. Fig. \ref{fig:chart} displays the performance enhancement of adversarial transfer (red color), AdvCont (yellow color), AdvCon with AutoAug (green color), and ALICE (orange color) compared with that of the baseline model (blue color) for different benchmark datasets. 
The margin difference accuracies of the proposed ALICE model compared with that of the standard baseline model for CIFAR100, CIFAR10, SVHN, Aircraft, Pets, and Nancho are enhanced by $3.4\%$, $2.9\%$, $3.0\%$, $3.4\%$, $2.1\%$, and $5.9\%$, respectively.

\begin{table}[ht!]
\caption{Performance of $AdvCont+AutoAug$ with various coefficient values of $\alpha_1$ (low-rank) and $\alpha_2$ (sparsity) to evaluate the impact of low-rank and sparsity constraints on CIFAR10 and Nancho datasets.}
\label{tbl:alpha_values}
\begin{center}
\begin{tabular}{|l|c|cccccc|}
\hline
\textbf{Dataset}    &\textbf{Constraint}    &   &   &\textbf{Range}   &  &  &    \\
   &   &\textbf{0}  &\textbf{0.3} &\textbf{0.5}  &\textbf{0.8}   &\textbf{1}  &\textbf{1.2}  \\
\hline \hline
\textbf{CIFAR10}    &$\boldsymbol{\alpha_1}$ &97.1     &97.4    &\textbf{97.7}   &97.5   &97.2  &97.0\\
                    &$\boldsymbol{\alpha_2}$       &97.1     &97.2     &97.2   &97.3   &\textbf{97.4}  &97.2 \\ \hline
\textbf{Nancho}     &$\boldsymbol{\alpha_1}$     &81.0     &81.4    &\textbf{81.8}   &81.3   &81.1  &80.8\\  
                    &$\boldsymbol{\alpha_2}$       &81.0     &81.1     &81.2   &81.4   &\textbf{81.5}  &81.3 \\    \hline
     
\end{tabular}
\end{center}
\end{table}

The hyperparameters in Eq. \eqref{eq:Lan_objfunc} for the unconstrained extended form of the augmented Lagrangian algorithm are obtained as $\alpha_1=0.5$, $\alpha_2=1$. The hyperparameters $\alpha_1$ and $\alpha_2$ are the coefficients that balance the impact of each low-rank and sparsity constraint along with the square-root penalty term. The values of $\alpha_1$ and $\alpha_2$ can have different impacts based on their weights. The performance of $AdvConv+AutoAug$ based on different values of $\alpha_1$ and $\alpha_2$ is presented in Table \ref{tbl:alpha_values} to evaluate the impact of low-rank and sparsity constraints on the CIFAR10 and Nancho datasets. Table \ref{tbl:alpha_values} shows that the most effective value for $\alpha_1$ for eliminating redundant information is $\alpha_1=0.5$ and values larger than $1$ are harmful to the model when a stronger low-rank constraint is applied. The impact of $\alpha_2$ on the sparsity constraint is also investigated, which shows that the value of $\alpha_2=1$ is the optimum sparse weight, and the larger values decrease the performance.

\paragraph{Computational Efficiency} In general, the targeted adversarial attacks such as the projected gradient descent with k update steps called PGD(k) is k times more expensive than natural training \citep{madry2019deep}. This is because there are k iterations in the inner maximization loop of the loss minimization objective function. However, as we utilized the pre-trained adversarially trained weights for transfer learning to downstream tasks, the computational cost of fine-tuning to a target dataset for both adversarial and standard models is the same. 
In the Lagrangian fine-tuning module, we utilize iSQRT bilinear transformation that uses Newton-Schulz iteration to approximate matrix square-root normalization, which requires only matrix multiplication. This is compatible with GPU acceleration. As the Lagrangian approach alternately optimizes the square-root, low-rank, and sparsity regularizers using only matrix multiplication, the time costs of all three regularizers are small, and the extra time consumed by updating the Lagrange multiplier is negligible.

\begin{figure}[ht!]
\centering\includegraphics[width=0.9\linewidth]{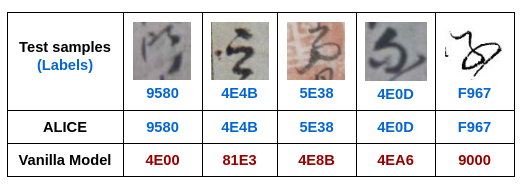}
\caption{Recognition comparison between the proposed ALICE and vanilla ResNet. }~\label{fig:lag_mis}
\end{figure}
 
Fig. \ref{fig:lag_mis} depicts the test samples with their corresponding class label (Uni-code) that are well-recognized using the proposed ALICE model while mis-recognized using vanilla model. The misclassified Uni-codes by the vanilla ResNet are shown in red. The vanilla model is not capable of proper recognition and mis-classifies the samples due to lack of discrimination. For example, in the second column, the test sample with Unicode $9580$ is mis-recognized as $4E00$, whereas ALICE correctly classifies it. 

\section{Conclusion}
\label{sec:conclusion}
We proposed a new AdvCont method to generate better representation and fine-tune the obtained representation with multi-objective Lagrangian multipliers to enable the low-rank and sparsity for small datasets. First, a higher representative embedding was acquired with the Min-Max adversarial loss. Then, the obtained embedding was utilized for the initialization of CL with shared weights. Adversarial CL resulted in higher accuracy and faster convergence for various subsets of datasets with a limited number of samples. The proposed method considered the labels using various augmentation techniques. Furthermore, augmented Lagrangian multipliers were enforced to encourage the low-rank and sparsity in the structure of the proposed model to adaptively estimate the coefficients of the regularizers. The low-rank module successfully removed trivial components of the model, and sparsity efficiently suppressed less representative feature elements. The performance of the proposed model was verified on benchmark datasets with limited data.

In future works, we intend to consider the orthogonality constraint in the context of the Lagrangian method to obtain better representations. Moreover, we plan to tweak the feature space by using within- and between-feature space regularizations.

\section*{Acknowledgment}
This work was partly supported by the National Research Foundation of Korea (NRF) grant funded by the Korean government (MSIT) (No. 2022R1A5A7026673) and also Electronics and Telecommunications Research Institute (ETRI) grant funded by the Korean government (22ZS1100, Core Technology Research for Self-Improving Integrated Artificial Intelligence System).

\bibliography{mybibfile}

\end{document}